\documentclass[lettersize,journal]{IEEEtran}
\usepackage{amsmath,amsfonts}
\usepackage{algorithmic}
\usepackage{algorithm}
\usepackage{array}
\usepackage[caption=false,font=normalsize,labelfont=sf,textfont=sf]{subfig}
\usepackage{textcomp}
\usepackage{stfloats}
\usepackage{url}
\usepackage{verbatim}
\usepackage{graphicx}
\usepackage{cite}
\usepackage{hyperref}
\usepackage{url}
\usepackage{booktabs}
\usepackage{bm}

\usepackage{color}
\usepackage{MnSymbol,wasysym}
\usepackage{multirow}
\usepackage{booktabs}
\usepackage{colortbl}
\usepackage[table]{xcolor}

\begin{document}

\title{SwimVG: Step-wise Multimodal Fusion and Adaption for Visual Grounding}


\author{Liangtao Shi, Ting Liu, 
Xiantao Hu, Yue Hu, Quanjun Yin, 
Richang Hong, Senior Member, IEEE
\thanks{\IEEEauthorrefmark{1} Liangtao Shi and Ting Liu contributed equally to this paper.}
\thanks{\IEEEauthorrefmark{2} Richang Hong is the corresponding author of this paper.}

\thanks{Liangtao Shi and Richang Hong with the Key Laboratory of Knowledge Engineering with Big Data, Hefei University of Technology, Hefei 230009, China, and also with the Ministry of Education and School of Computer Science and Information Engineering, Hefei University of Technology, Hefei 230009, China (e-mail: shilt@mail.hfut.edu.cn;
hongrc.hfut@gmail.com).}

\thanks{Ting Liu, Yue Hu and Quanjun Yin are with School of systems engineering, National University of Defense Technology, Changsha, Hunan Province, 410073, China. (e-mail: liuting20@nudt.edu.cn; yquanjun@126.com; huyue11@nudt.edu.cn). Xiantao Hu with the Department of Computer Science and Engineering, Nanjing University of Science and Technology, Nanjing 210014, China (e-mail: huxiantao481@gmail.com). This research was partially supported by the National Natural Science Fund of China (Grant Nos. 62306329 and 62103425), Natural Science Fund of Hunan Province (Grant Nos. 2023JJ40676 and 2022JJ40559).}  

\thanks{}
\thanks{}
\thanks{}
}

\markboth{Journal of \LaTeX\ Class Files,~Vol.~14, No.~8, August~2021}%
{Shell \MakeLowercase{\textit{et al.}}: A Sample Article Using IEEEtran.cls for IEEE Journals}


\maketitle

\begin{abstract}
\label{sec:abstract}
Visual grounding aims to ground an image region through natural language, which heavily relies on cross-modal alignment. Most existing methods transfer visual/linguistic knowledge separately by fully fine-tuning uni-modal pre-trained models, followed by a simple stack of visual-language transformers for multimodal fusion. However, these approaches not only limit adequate interaction between visual and linguistic contexts, but also incur significant computational costs. Therefore, to address these issues, we explore a step-wise multimodal fusion and adaption framework, namely SwimVG. Specifically, SwimVG proposes step-wise multimodal prompts (Swip) and cross-modal interactive adapters (CIA) for visual grounding, replacing the cumbersome transformer stacks for multimodal fusion. Swip can improve {the} alignment between the vision and language representations step by step, in a token-level fusion manner. In addition, weight-level CIA further {promotes} multimodal fusion by cross-modal interaction. Swip and CIA are both parameter-efficient paradigms, and they fuse the cross-modal features from shallow to deep layers gradually. Experimental results on four widely-used benchmarks demonstrate that SwimVG achieves remarkable abilities and considerable benefits in terms of efficiency. Our code is available at \url{https://github.com/liuting20/SwimVG}.

\end{abstract}
\section{Introduction}

Visual grounding (VG) \cite{kamath2021mdetr,vg-law,qiao2020referring,xiao2024towards} refers to locating the bounding box region described by a textual expression in a specific image, which is one of the most challenging tasks in multimodal fields. In contrast to vanilla detection tasks, VG requires fine-grained vision-language alignment so as to precisely locate an object described through a language expression. The evolution of VG has considerable potential to promote vision-language understanding, and enjoys broad applications in fields such as robot navigation\cite{das2018embodied}, visual Q\&A \cite{antol2015vqa} and automatic driving\cite{zhang2022ri,motroni2020sensor}.

\begin{figure}[t]
\centering
\includegraphics[width=1.0\linewidth]{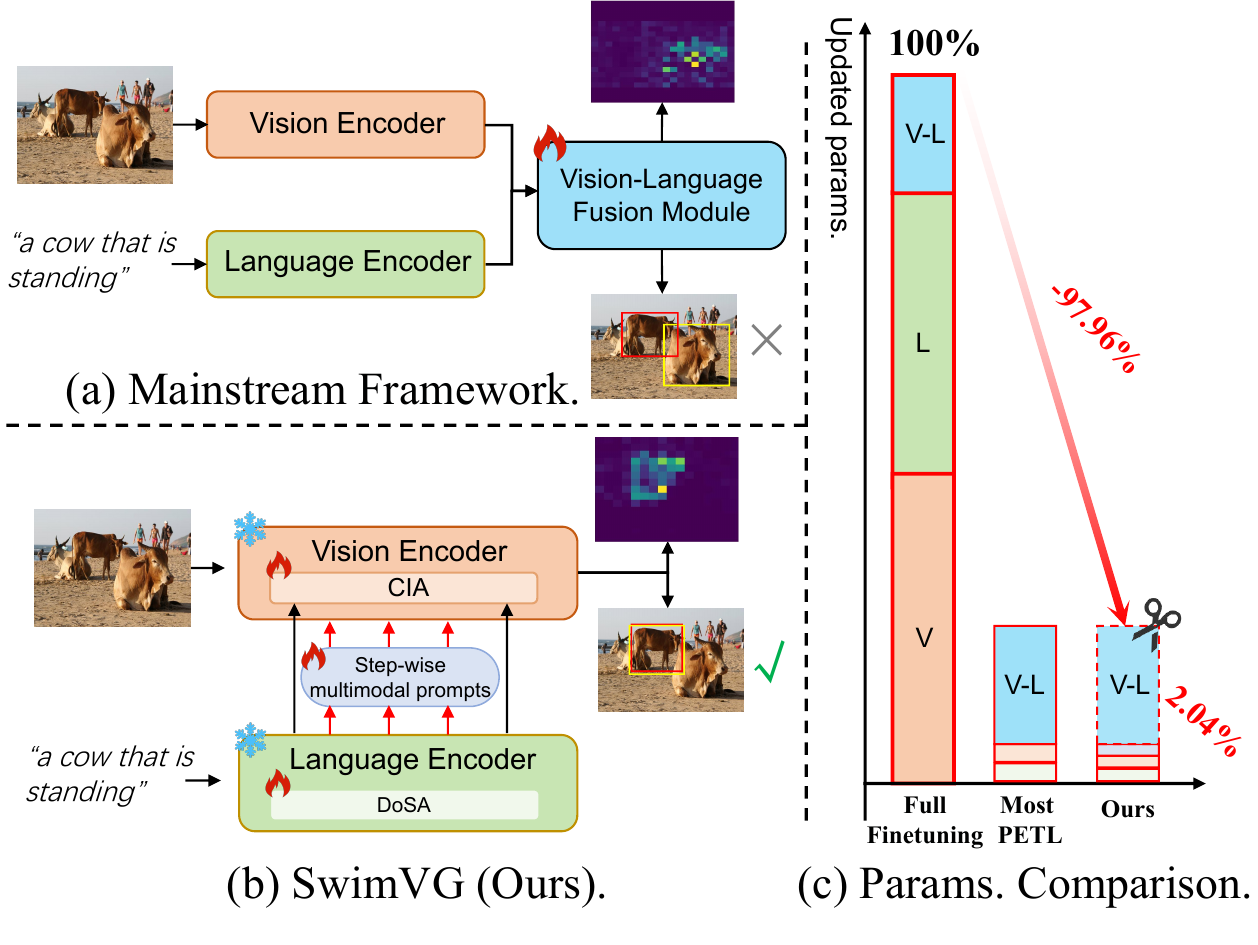}
\caption{Comparison of multimodal fusion strategy between (a) mainstream {framework} and (b) SwimVG (ours) for visual grounding. Freezing the pre-trained models (\protect\includegraphics[height=0.5cm]{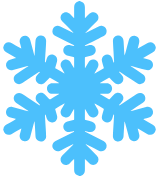}) and only updating (\protect\includegraphics[height=0.5cm]{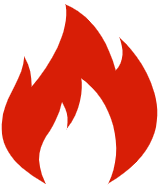}) the tiny modules in SwimVG reduces 97.96\% updated parameters while achieving even stronger performance.}
\label{fig:intro}
\end{figure}

Early visual grounding methods followed target detection frameworks, evolving from the initial two-stage approaches to the recent one-stage methods. Benefiting from the open source of transformer-based pre-trained models,  
a growing number of approaches \cite{deng2021transvg,kamath2021mdetr,transvg++,shi2023dynamic} transfer the language and vision knowledge from pre-trained models by fully fine-tuning, such as TransVG \cite{deng2021transvg}, TransVG++ \cite{transvg++}, and VG-LAW \cite{vg-law}. These methods commonly adopt visual and textual encoders to extract {features}, respectively, which are subsequently input into a vision-language (VL) transformer for cross-modal interaction. As shown in Fig.\ref{fig:intro}(a), we visualize the last layer of vision-language transformer in mainstream method, it indicates that the visual attentions focus on the foreground area of the image, rather than the text-relevant region (``$standing$''). We have summarized two reasons for this phenomenon:

\begin{itemize}
  \item The vision-language encoder for multimodal fusion is a coarse stack of transformers, the mechanism not only limits the sufficient interaction between vision and language contexts, but also exacerbates the computational cost due to the deep transformer-based structure.
  \item 
  Fine-tuning the entire backbone might suffer catastrophic forgetting and undermine the extensive prior knowledge learned from pre-training. In addition, fully training large pre-trained models and the VL transformer can be computationally expensive and time-consuming in practice.

\end{itemize}

Several previous works have noticed the insufficient
interaction problem, QRNet \cite{ye2022shifting} achieves the expression-aware visual feature extraction by inserting carefully designed interaction modules into the visual backbone. VG-LAW \cite{vg-law} proposes a language adaptive weight generator, generating weights for the fusion of visual and text features. However, they still require fully fine-tuning backbone and the sophisticated designs of interactive modules. More recently, Parameter-Efficient Transfer Learning (PETL) methods have also been introduced into visual grouding \cite{xiao2024hivg,liu2024dara}, HiVG \cite{xiao2024hivg} adopts LoRA to fine-tune the frozen CLIP model, and DARA\cite{liu2024dara} designs DA adapter and RA adapter to transfer intra- and inter-modality representations for the VG domain. However, due to the simple fusion strategy of vision-language transformer, they are not sufficient for multimodal alignment,  which could potentially compromise the model’s ability to capture text-relevant visual details.

In this paper, we aim to explore an efficient tuning and lightweight cross-modal interaction strategy. Inspired by the efficiency of Prompt Tuning \cite{khattak2023maple,liu2024dap,shi2024explicit} and Adapter \cite{chen2022adaptformer}, which only require fine-tuning a tiny number of parameters to adapt pre-trained models to various downstream tasks. We propose a step-wise multimodal fusion and adaption framework (SwimVG). {As depicted in Fig. \ref{fig:intro}(b), we design step-wise multimodal prompts (Swip) for multimodal fusion step by step, and explore a cross-modal interactive adapter (CIA) for further vision-text alignment. The visualizations of Swip (Fig. \ref{fig:more visualizations}(b)) and CIA (Fig. \ref{fig:more visualizations}(c)) demonstrate that both of them can independently facilitate multimodal interaction. Their integration, namely SwimVG, as visualized in Fig. \ref{fig:more visualizations}(c), leads to enhanced multimodal fusion.}
Through these elaborate designs, we implement an efficient and effective multimodal fusion strategy, abandoning the additional vision-language transformers used in previous methods \cite{deng2021transvg,transvg++,xiao2023clip,liu2024dara,xiao2024hivg}. As shown in Fig. \ref{fig:overview} (b), the vision attentions of the last layer in vision encoder indicate that SwimVG focuses exactly the text-relevant region.
 
Specifically, to efficient tuning the whole network, we frozen the vision and text backbone, and adopt domain-specific adapters (DoSA) for transferring pre-trained language knowledge to the specific task. To achieve adequate multimodal alignment, we investigate two strategies, namely token-level and weight-level. {For token-level multimodal fusion, we design step-wise multimodal prompts, which is formed by gradually integrating a learnable token that can represent the global text semantics into the visual backbone layer by layer.} These tokens are initially placed on the language encoder layers, and then mapped to the visual encoder from shallow to deep layers. To further enhance the multimodal fusion in a weight-level manner, we propose a novel cross-modal interactive adapter, which integrate visual and textual features by multi-head cross-attention mechanism. The multimodal adaptation process involves a set of low-rank weight matrices reorganized, producing the crucial alignment capabilities for visual grounding. By the multi-level design of token- and weight-level, for a given image input, the visual encoder can focus more on the text-relevant area, without fully fine-tuning the pre-trained models. 

We conduct extensive experiments on RefCOCO \cite{yu2016refcoco}, RefCOCO+ \cite{yu2016refcoco}, RefCOCOg \cite{mao2016refcocogg,nagaraja2016refcocogu} and Flickr30K Entities \cite{plummer2015flickr30k}, and our method achieves state-of-the-art (SOTA) performance on the four widely used datasets. In addition, we demonstrate the efficiency of our framework in Table \ref{tab:cost}, it can be seen that the inference time of SwimVG is about 40\% faster than these mainstream methods using the vision-language transformer. The main contributions can be summarized as three-fold:
\begin{itemize}
  \item 
  We proposed a concise and efficient framework of multimodal fusion and adaption, which adapt pre-trained models to visual grounding step by step. SwimVG achieves token-level and weight-level interaction between visual and language representations, and significantly alleviates the task gap between pre-trained models and grounding.
  
  \item We replace the heavyweight vision-language transformer with cross-modal interactive adapters and step-wise multimodal prompts, which allow for fine-tuning the entire model in a lightweight manner.

  \item
  Extensive experiments demonstrate that our method outperforms the SOTA methods in VG tasks, with only \textbf{2.04\%} tunable parameters. Moreover, SwimVG offers significant computing efficiency advantages.

\end{itemize}

\section{Related Work}
\label{sec:related work}

\subsection{Visual Grounding}
Visual grounding (VG)~\cite{yu2018mattnet,yang2019fast,deng2021transvg,xiao2023clip,liu2024dara,xiao2024hivg,lu2024lgr} aims to identify and localize regions within images that correspond to given text descriptions. 
There are many extensions of VG in other fields, such as Remote Sensing VG \cite{li2024language,li2024show,ding2024visual}.
Early visual grounding methods, given their resemblance to detection tasks, initially aligned with the prevailing object detection architectures. These architectures evolve from the initial two-stage designs to the recently one-stage methods. Two-stage designs methods~\cite{liu2019learning,hong2019learning,chen2020referring} follow a two-stage pipeline that first utilizes pre-trained object detectors to obtain a set of region proposals, which are then ranked based on their similarity scores with the given textual description. However, these two-stage methods face challenges in terms of the performance of the proposal generators and the additional ranking mechanisms. With the introduction of ViT, the Transformer-based methods~\cite{deng2021transvg,du2022vgtr,yang2022vltvg,zhu2022seqtr,su2023referring,vg-law,liu2024dara,zhu2023jmri} further propose an end-to-end framework which reformulate the prediction process as a regression problem. Most recently, grounding multimodal large language models~\cite{bai2023qwen,wang2024visionllm,wang2023cogvlm} have propelled the state-of-the-art (SOTA) performance, these works require a large amounts of in-domain and other domain datasets. Despite the transformer-based models exhibiting ideal performance in VG, most methods involve fully fine-tuning the text and visual branches separately, followed by a heavyweight vision-language encoder for simple multimodal fusion. This not only makes it difficult to focus on the areas most relevant to the text description but is also inefficient.

\subsection{Parameter-Efficient Transfer Learning}

Transfer learning aims to adapt pre-trained models to specific tasks or datasets. With the growth of model sizes and the complexity of the specific tasks, fully fine-tuning paradigm demands significant computational resources. To address these challenges, researchers in the NLP and CV domains have explored PETL methods ~\cite{lester2021power,hu2021lora,chen2022adaptformer,yuan2023mrsadapter,zhou2024dynamic}. One method, known as Prompt Tuning \cite{wang2023aprompt,khattak2023maple,liu2024dap}, involves the introduction of trainable tokens at the input space, thereby learning task-specific representations. Adapter-like methods \cite{yuan2023mrsadapter,liu2024sparse} involve inserting additional trainable weights, such as Multi-Layer Perceptrons (MLPs) equipped with activation functions and residual connections, within the network architecture to enhance transfer learning capabilities. Meanwhile, LoRA-like methods \cite{hu2021lora} adjust pre-trained models by using the idea of low-rank matrix decomposition, and only trains the parameters of the low-rank matrix. LoRA-like methods have been proposed in the field of natural language processing for Large Language Models (LLM) such as GPT-4 \cite{achiam2023gpt}, LLaMA2 \cite{touvron2023llama}, and GLM-4 \cite{glm2024chatglm}. By focusing on updating only a small subset of parameters, PETL methods effectively simulate the fine-tuning of the entire model’s parameters without directly modifying them. Recently, some pioneering works like MaPPER~\cite{liu2024mapper}, HiVG\cite{xiao2024hivg}, DARA~\cite{liu2024dara} and M$^2$IST \cite{liu2024m} sought to utilize adapters to adapt pre-trained models to visual grounding. However, they all use a burdensome vision-language module for multimodal fusion, which is not an efficient enough method.

\begin{figure*}[t!]
\centering
\includegraphics[width=1.0\textwidth]{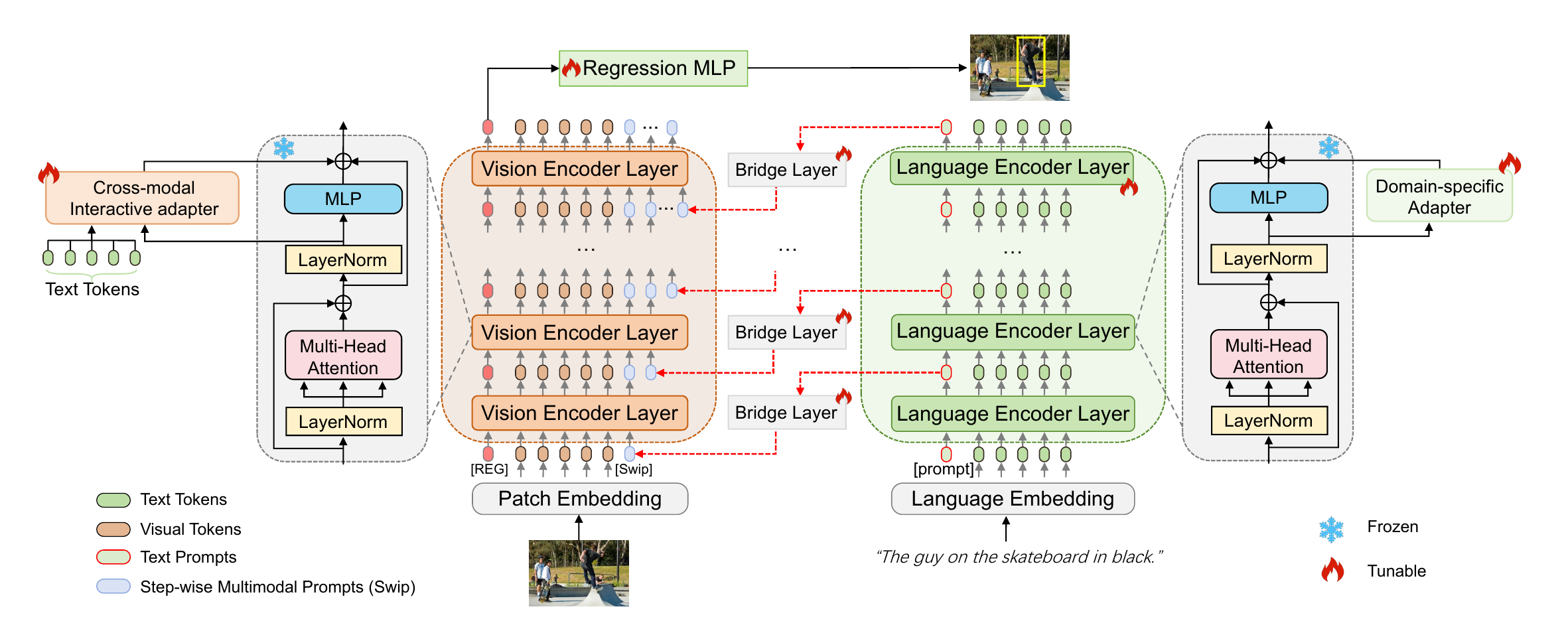}
\caption{Overall architecture of the proposed SwinVG, which freezes the pre-trained vision encoder and language encoder. SwimVG integrates step-wise multimodal prompts (Swip) and cross-modal interactive adapters, which bridges the visual and language encoders, ensuring the visual encoder concentrates on the text-relevant areas.}
\label{fig:overview}
\end{figure*}

\section{Method}

Our method is designed to enhance the generalization capabilities of pre-trained models in the realm of visual grounding efficiently. This is achieved through step-wise multimodal prompts, light domain-specific adapters, and cross-modal interactive adapters. Fig. \ref{fig:overview} shows the overall architecture of our proposed SwimVG framework.

\subsection{Text \& Image Backbone}
Given an image and a text, we extract their features through the image encoder and text encoder, respectively.

\noindent
\textbf{Text Encoder.} Given the input text expression $T$ with a length of $L$, we utilize the pre-trained text branch of CLIP \cite{radford2021learning} for extracting text features. The text expression is firstly converted into a one-hot vector. Subsequently, each one-hot vector is tokenized into a series of linguistic tokens, and the sequence of tokens is then fed into a stack of 12 transformer encoder layers to progressively capture and model the intricate language tokens.
The input embeddings $\bm{\hat T} \in \mathbb{R}^{L\times C_t}$, where $\bm{\hat T} = [t^1, t^2, \cdots, t^L]$, and $C_t$ is the dimension of text embeddings.

\label{text encoder:clip}

\noindent
\textbf{Visual Encoder.} 
We use DINOv2 \cite{oquab2023dinov2} as the visual backbone. The model involves training the Vision Transformer (ViT) model \cite{dosovitskiy2020vit} on the extensive LVD-142M dataset, and employs a self-supervised learning strategy. This method allows the model to extract powerful visual features, thereby offering remarkable performance in various downstream tasks. Given an input image
${I}_0 \in \mathbb{R}^{H_0\times W_0\times3}$, the image is initially divided into $N$ non-overlapping patches, which are then linearly projected into $D$-dim patch embeddings $\bm{I'}_0 \in \mathbb{R}^{D\times C_v}$. Motivated by TransVG \cite{deng2021transvg} and HiVG \cite{xiao2024hivg} appending a learnable [REG] in vision-language transformer, we also adopt a learnable [REG] token to directly predict the 4-dim coordinates of a referred object. Unlike the previous method, we omit the complex vision-language fusion structure and directly pre-append the [REG] token to $\bm{I'}_0$, and the token is processed by the visual encoder layer gradually.

As the vision and language backbones contain most model parameters and have acquired rich knowledge during pre-training, we attempt to freeze them during fine-tuning. This strategy allows for a more efficient allocation of computational resources and focuses the learning on the adjustments of other modules.

\subsection{Step-wise Multimodal Prompting} 
The intuitive idea of achieving token-level multimodal alignment is to directly concatenate text tokens and vision tokens together for learning. However, an increase in the input length will bring about a computational burden. To efficiently establish token-level multimodal alignment, we design step-wise multimodal prompts, and introduce these learnable tokens in the layers of both vision and language branches from shallow to deep layer. This means that these tokens are added to transformer layers in a hierarchical way. The hierarchical multimodal prompts utilize the knowledge embedded in pre-trained models to effectively learn task-relevant cross-modal representations. 


\begin{figure}
\centering
\includegraphics[width=1.0\linewidth]{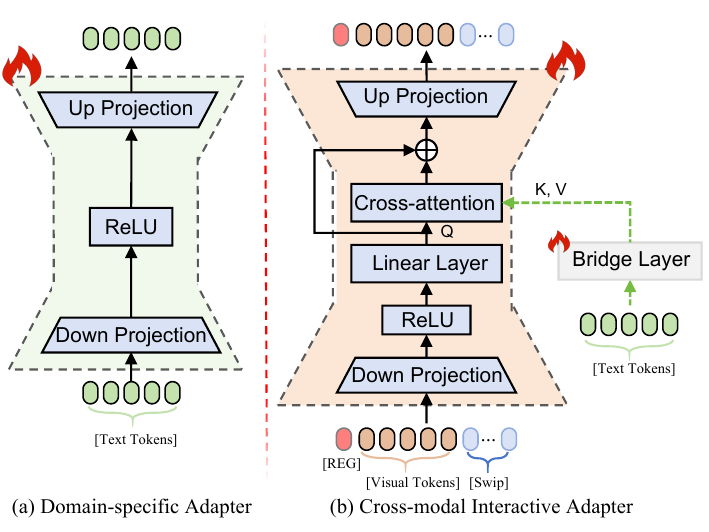}
\caption{The Domian-specific adapter and cross-modal interactivate adapter.}
\label{fig:adapter}
\end{figure}

\noindent
\textbf{Text Prompting.}
To learn the represent the global textual representation, a learnable token $p \in \mathbb{R}^{H}$ is introduced in the text encoder. The input embeddings $\bm {\hat T}$ is converted to $ \bm{\hat T'}$, follow the form $[p,t^1, t^2, \cdots, t^L]$. The new token is further processed by each transformer block of the language encoder $\mathcal{L}_{i}$. This process can be formulated as below:

\begin{equation}
\begin{aligned}
[{p_i}, \bm {T}_{i}] = \mathcal{L}_{i}([p_{i-1}, \bm {T}_{i-1}]) ~~~~ i=1, 2, \cdots, l.
\end{aligned}
\end{equation}

\noindent
where $p_i$ and \bm ${T}_{i}$ represent the prompt and text embeddings processed by the $i$-th language encoder layer, respectively. The $l$ refers to the depth of the language encoder. The prompt is initialized by Xavier initialization.

\noindent
\textbf{Step-wise Multimodal Fusion.}
To efficiently fuse textual and visual semantics step-by-step, we gradually convey the prompt processed by the text encoder to vision encoder.
Due to the different feature dimensions of the text and vision encoder, we need to adjust the features to the same dimension. Therefore, we design a bridge layer to transport text features, making them adaptable for visual branch. For the visual encoder layer $\mathcal{V}_{i-1}$, we introduce the token from the $\mathcal{L}_{i}$ language encoder layer to the layer $\mathcal{V}_{i-1}$ of vision encoder. Since the prompt added to the visual token set is initialized by global textual semantics, when the prompt is introduced into the corresponding visual layer, it can facilitate multimodal fusion step by step, namely step-wise multimodal prompt (Swip). Each Swip is further processed by the deeper visual layer. The process can be formalized as:

\begin{equation}
m_i = p_i\mathbf{W}_{bridge}
\end{equation}

\noindent

\begin{equation}
\begin{aligned}
[m_{i}^0,\cdots,m_{i}^{i-1}, \bm V_{i}] = \mathcal{V}_{i}([m_{i-1}^0,\cdots,m_{i-1}^{i-1},\bm V_{i-1}]) \\
\end{aligned}
\end{equation}

\noindent
where $\mathbf{W}_{\text{bridge}} \in \mathbb{R}^{C_t \times C_v}$ is the weights of bridge layer. The $m_i$ are multimodal prompts transformed from text prompts. The $n$ refers to the depth of vision encoder, and $m_i^{0}$ represents the $0$-th swip token processed by the $i$-th vision encoder layer. The $\bm V_{i}$ is vision emdeddings processed by the $i$-th vision encoder layer.

\subsection{Cross-modal Interactive \& Domain Adaption}

To efficiently transfer pre-trained text semantics knowledge to visual grounding, and further facilitate multimodal interaction, we introduce domain-specific adapters in the text encoder and cross-modal interactive adapters in vision encoder, respectively.

\noindent
\textbf{Cross-modal Interactive Adapter.}
As shown in Fig. \ref{fig:adapter}(b), we design a Cross-modal Interactive Adapter (CIA) to make the interaction of modal information between the visual encoder and text encoder, which enhances the capability of multimodal fusion while fixing the backbone parameters. The main difference between the design of CIA and previous adapters lies in the integration of a cross-modal attention module. To ensure the lightweight and efficiency of whole structure, we firstly adopt a down-projection to transform the visual features to low-rank features. CIA module is inserted between the activation and up-projection layers. Similar to step-wise multimodal fusion, text features should be converted by bridge layer to dimensions that match the visual branches. Given vision features $f_i^v$ process by the Multi-Head Attention (MHA) of the layer $\mathcal{V}_{i}$, and text features $f_i^t$ process by the MHA of the layer $\mathcal{L}_{i}$, this process can be formulated as below:

\begin{equation}
c_i^t = f_t\mathbf{W}_{bridge}
\end{equation}

\begin{equation}
\begin{aligned}
\begin{split}
    f_{down}^v = f_i^v\mathbf{W}_{down}, \\
    f_{act}^v=\operatorname{ReLU}\boldsymbol(f_{down}), \\
    f_{l}^v = f_{act}\mathbf{W}_{linear}.
\end{split}
\end{aligned}
\end{equation}

\begin{equation}
\small
\begin{aligned}
\mathbf{MHCA}(f_{l}^v,c_i^t)=\operatorname{Softmax}\left(\frac{f_i^vW_qc_i^tW_k}{\sqrt{C}}\right)(c_i^tW_v)
\end{aligned}
\end{equation}

\begin{equation}
f_{up} = (f_{l}^v+\mathbf{MHCA}(f_{l}^v,c_i^t))\mathbf{W}_{up}
\end{equation}

\begin{equation}
 \mathbf{CIA}(f_i^v,c_i^t)= f_i^v + s_{vt} \cdot f_{up}.
\end{equation}

\noindent
Here, $\mathbf{W}_{\text{down}} \in \mathbb{R}^{C_v \times C_d}$ and $\mathbf{W}_{\text{up}} \in \mathbb{R}^{C_d \times C_v}$ are the weights of down- and up-projection layers, and $s_{vt}$ is the scaling factor for multimodal fusion. The $\mathbf{MHCA}$ is Multi-Head Cross-Attention module in CIA.

\noindent
\textbf{Domain-specific Adapter.}
Due to fully freezing of the text backbone, there exists a gap between pre-trained model and visual grounding. To address the issue, we incorporate domain-specific adapters (DoSA) to improve the text encoder for domain understanding, As shown in Fig. \ref{fig:adapter}(a). Compared to the CIA adapter, the domain-specific adapter adopts a more straightforward design, focusing on learning text representation efficiently without complex structure. This neat yet effective approach ensures efficient processing of text semantics while maintaining compatibility with the overall model architecture. By taking advantage of these enhanced features, the model facilitates aligning visual and linguistic features. Specifically, the domain-specific adapter follows a standard ``Down-ReLU-Up" structure to bridge the gap between pre-trained knowledge and visual grounding. Given the text features $f_i^t$ processed by the Multi-Head Attention (MHA) of the layer $\mathcal{L}_{i}$, the learning process can be formalized as:

\begin{table*}[!t]
\centering
\caption{Comparison with latest SOTA methods on RefCOCO/+/g for visual grounding. "RN50", "RN101", "DN53", and "Swin-S" represent ResNet-50 \cite{he2016resnet}, ResNet-101 \cite{he2016resnet}, DarkNet-53 \cite{redmon2018yolov3}, and Swin-Transformer Small, respectively. $\dagger$ indicates that all of the RefCOCO/+/g training data has been used during pre-training. ``Tuned/Total param.'' is the average percentage of tuned parameters in whole model. The boldface denotes the best performance while the underline indicates the second best. }
\label{Table:comparisons with SOTA}
\setlength{\tabcolsep}{2.2pt}
\resizebox{1\textwidth}{!}{
\begin{tabular}{lllccccccccccll}
\toprule

\multicolumn{1}{c|}{\multirow{2}{*}{Methods}}&\multicolumn{1}{c|}{\multirow{2}{*}{Venue}}&\multicolumn{1}{c|}{\multirow{2}{*}{Backbone}}
& \multicolumn{1}{c|}{Tuned/Total} & \multicolumn{3}{c|}{RefCOCO} & \multicolumn{3}{c|}{RefCOCO+} & \multicolumn{3}{c|}{RefCOCOg} &{Flickr30K} & \\

  \multicolumn{1}{c|}{}& \multicolumn{1}{c|}{}&\multicolumn{1}{c|}{}& \multicolumn{1}{c|}{param.} & val & testA & \multicolumn{1}{c|}{testB} & val & testA & \multicolumn{1}{c|}{testB} & val-g & val-u & \multicolumn{1}{c|}{test-u} & \multicolumn{1}{c}{test}& \\ \midrule

\rowcolor{gray!20}\textbf{Full Fine-tuning}  & & & &    &  &  &  && &  & &   & &\\
\midrule

\multicolumn{1}{c|}{MAttNet~\cite{yu2018mattnet}}    &\multicolumn{1}{c|}{CVPR'18} &\multicolumn{1}{c|}{RN101/LSTM}& \multicolumn{1}{c|}{100\%} & 76.65 & 81.14 &  \multicolumn{1}{c|}{69.99} & 65.33 & 71.62 & \multicolumn{1}{c|}{56.02} & - & 66.58 & \multicolumn{1}{c|}{67.27}  & \multicolumn{1}{c}{-}&\\

\multicolumn{1}{c|}{RvG-Tree~\cite{hong2019learning}}   &\multicolumn{1}{c|}{TPAMI'19} &\multicolumn{1}{c|}{RN101/LSTM}& \multicolumn{1}{c|}{100\%} & 75.06 & 78.61 &  \multicolumn{1}{c|}{69.85} & 63.51 & 67.45 & \multicolumn{1}{c|}{56.66} & - & 66.95 & \multicolumn{1}{c|}{66.51}  & \multicolumn{1}{c}{-}&\\

\multicolumn{1}{c|}{NMTree~\cite{liu2019learning}}   &\multicolumn{1}{c|}{ICCV'19}& \multicolumn{1}{c|}{RN101/LSTM}& \multicolumn{1}{c|}{100\%} & 76.41 & 81.21 &  \multicolumn{1}{c|}{70.09} & 66.46 & 72.02 & \multicolumn{1}{c|}{57.52} & 64.62 & 65.87 & \multicolumn{1}{c|}{66.44}  &\multicolumn{1}{c}{-} &\\ 

\multicolumn{1}{c|}{FAOA~\cite{yang2019fast}}  &\multicolumn{1}{c|}{ICCV'19} &\multicolumn{1}{c|}{DN53/LSTM}& \multicolumn{1}{c|}{100\%} & 72.54 & 74.35 &  \multicolumn{1}{c|}{68.50} & 56.81 & 60.23 & \multicolumn{1}{c|}{49.60} & 56.12 & 61.33 & \multicolumn{1}{c|}{60.26}  &\multicolumn{1}{c}{68.71} &\\

\multicolumn{1}{c|}{ReSC-Large~\cite{yang2020improving}}   &\multicolumn{1}{c|}{ECCV'20} &\multicolumn{1}{c|}{ND53/BERT-B}& \multicolumn{1}{c|}{100\%} & 77.63 & 80.45 & \multicolumn{1}{c|}{72.30} & 63.59 & 68.36 & \multicolumn{1}{c|}{56.81} & 63.12 & 67.30 & \multicolumn{1}{c|}{67.20}  &\multicolumn{1}{c}{69.28} &\\ 

\multicolumn{1}{c|}{TransVG~\cite{deng2021transvg}}   & \multicolumn{1}{c|}{ICCV'21}& \multicolumn{1}{c|}{RN50+DETR/BERT-B}& \multicolumn{1}{c|}{100\%}&  80.32 & 82.67 & \multicolumn{1}{c|}{78.12} & 63.50 & 68.15 & \multicolumn{1}{c|}{55.63} & 66.56 & 67.66 & \multicolumn{1}{c|}{67.44}  &\multicolumn{1}{c}{78.47} &\\
\multicolumn{1}{c|}{QRNet~\cite{ye2022shifting}}     & \multicolumn{1}{c|}{CVPR'22}&\multicolumn{1}{c|}{Swin-S/BERT-B}& \multicolumn{1}{c|}{100\%}   & 84.01  & 85.85  & \multicolumn{1}{c|}{82.34}  & 72.94  & 76.17  & \multicolumn{1}{c|}{63.81}  & 71.89  & 73.03  &\multicolumn{1}{c|}{72.52}  &\multicolumn{1}{c}{81.95} &\\

\multicolumn{1}{c|}{Dynamic-MDETR $^\dagger$~\cite{shi2023dynamic}}  & \multicolumn{1}{c|}{TPAMI’23}& \multicolumn{1}{c|}{CLIP-B}&\multicolumn{1}{c|}{100\%}  & 85.97  & \underline{88.82}  & \multicolumn{1}{c|}{80.12}  & 74.83  & 81.70  & \multicolumn{1}{c|}{63.44}  & 72.21& 74.14  & \multicolumn{1}{c|}{74.49}   & \multicolumn{1}{c}{81.89} &\\

\multicolumn{1}{c|}{PFOS~\cite{sun2022proposal}}   &\multicolumn{1}{c|}{TMM’22}& \multicolumn{1}{c|}{DN53/BERT-B}& \multicolumn{1}{c|}{100\%} & 77.37 & 80.43 & \multicolumn{1}{c|}{72.87} & 63.74 & 68.54 & \multicolumn{1}{c|}{55.84} & 61.46 & 67.08 & \multicolumn{1}{c|}{66.35}  &\multicolumn{1}{c}{-} &\\
\multicolumn{1}{c|}{SeqTR~\cite{zhu2022seqtr}}   & \multicolumn{1}{c|}{ECCV'22} &\multicolumn{1}{c|}{DN5/BiGRU3}& \multicolumn{1}{c|}{100\%}   & 81.23  & 85.00  & \multicolumn{1}{c|}{76.08}  & 68.82  & 75.37  & \multicolumn{1}{c|}{58.78} & - & 71.35  & \multicolumn{1}{c|}{71.58}    &\multicolumn{1}{c}{81.23} &\\  
\multicolumn{1}{c|}{Word2Pix~\cite{zhao2022word2pix}}     & \multicolumn{1}{c|}{TNNLS’22} &\multicolumn{1}{c|}{RN101+DETR/BERT-B}& \multicolumn{1}{c|}{100\%}   & 81.20  & 84.39  & \multicolumn{1}{c|}{78.12}  & 69.46  & 76.81  & \multicolumn{1}{c|}{61.57} & -& 70.81  & \multicolumn{1}{c|}{71.34}      &- &\\
\multicolumn{1}{c|}{YORO$^\dagger$~\cite{ho2023yoro}}   & \multicolumn{1}{c|}{ECCV'22} &\multicolumn{1}{c|}{ViLT}& \multicolumn{1}{c|}{100\%}  & 82.90  & 85.60  & \multicolumn{1}{c|}{77.40}  & 73.50  & 78.60  & \multicolumn{1}{c|}{64.90} & - & 73.40  & \multicolumn{1}{c|}{74.30}    &\multicolumn{1}{c}{-} &\\

\multicolumn{1}{c|}{TransVG++~\cite{transvg++}}  & \multicolumn{1}{c|}{TPAMI'23} &\multicolumn{1}{c|}{ViT-Det/BERT-B}& \multicolumn{1}{c|}{100\%} & \underline{86.28}  & 88.37  & \multicolumn{1}{c|}{80.97}  & 75.39  & 80.45  & \multicolumn{1}{c|}{66.28} &  73.86& 76.18  & \multicolumn{1}{c|}{76.30}  & \multicolumn{1}{c}{81.49}&\\ 

\multicolumn{1}{c|}{CLIP-VG~\cite{xiao2023clip}}  & \multicolumn{1}{c|}{TMM'23}& \multicolumn{1}{c|}{CLIP-B}& \multicolumn{1}{c|}{100\%}   & 84.29  & 87.76  & \multicolumn{1}{c|}{78.43}  & 69.55  & 77.33  & \multicolumn{1}{c|}{57.62} & 72.64 & 73.18  & \multicolumn{1}{c|}{72.54}  &\multicolumn{1}{c}{81.99} &\\ 

\multicolumn{1}{c|}{JMRI~\cite{zhu2023jmri}}  & \multicolumn{1}{c|}{TIM'23}& \multicolumn{1}{c|}{CLIP-B}& \multicolumn{1}{c|}{100\%}    & 82.97 & 87.30  & \multicolumn{1}{c|}{74.62}  & 71.17  & 79.82  & \multicolumn{1}{c|}{57.01} & 69.32  & 71.96  & \multicolumn{1}{c|}{72.04}   & \multicolumn{1}{c}{79.90}&\\

\multicolumn{1}{c|}{PTP2R-BLIP~\cite{wang2023enhancing}} & \multicolumn{1}{c|}{TPAMI'23}& \multicolumn{1}{c|}{BLIP} &\multicolumn{1}{c|}{100\%}   & 81.83  & 86.44  & \multicolumn{1}{c|}{74.30}  & \underline{76.65}  & \underline{82.14}  & \multicolumn{1}{c|}{\underline{67.38}} & - & -   & \multicolumn{1}{c|}{-}  &\multicolumn{1}{c}{-} &\\

\multicolumn{1}{c|}{VG-LAW~\cite{vg-law}}      & \multicolumn{1}{c|}{CVPR'23} &\multicolumn{1}{c|}{ViT-Det/BERT-B} &\multicolumn{1}{c|}{100\%} & 86.06  & 88.56  & \multicolumn{1}{c|}{\underline{82.87}}  & 75.74  & 80.32  & \multicolumn{1}{c|}{66.69} & - & 75.31   & \multicolumn{1}{c|}{75.95}     &\multicolumn{1}{c}{-} &\\

\multicolumn{1}{c|}{MGCross~\cite{miao2023self}}      & \multicolumn{1}{c|}{TIP'24}&\multicolumn{1}{c|}{RN101/BERT-B} & \multicolumn{1}{c|}{100\%}   & 85.10  & 88.23  & \multicolumn{1}{c|}{80.08}  & 74.44  & 79.48  & \multicolumn{1}{c|}{65.21} & 74.50 & \underline{77.25}   & \multicolumn{1}{c|}{75.78}   & \multicolumn{1}{c}{75.18}&\\

\multicolumn{1}{c|}{TransCP~\cite{tang2023context}}      & \multicolumn{1}{c|}{TPAMI'24} &\multicolumn{1}{c|}{RN50/BERT-B}& \multicolumn{1}{c|}{100\%}   & 84.25  & 87.38  & \multicolumn{1}{c|}{79.78}  & 73.07  & 78.05  & \multicolumn{1}{c|}{63.35} & 72.60 & -   & \multicolumn{1}{c|}{-}     &\multicolumn{1}{c}{80.04} &\\

\multicolumn{1}{c|}{LGR-NET~\cite{lu2024lgr}}      & \multicolumn{1}{c|}{TCSVT'24} &\multicolumn{1}{c|}{Swin-S/BERT-B}& \multicolumn{1}{c|}{100\%}   & 85.63  & 88.24  & \multicolumn{1}{c|}{82.69}  & 75.32  & 80.60  & \multicolumn{1}{c|}{68.30} & \underline{75.48} & 76.82   & \multicolumn{1}{c|}{\underline{77.03}} &\multicolumn{1}{c}{\underline{81.97}} \\

\multicolumn{1}{c|}{ScanFormer~\cite{su2024scanformer}}      & \multicolumn{1}{c|}{CVPR'24} &\multicolumn{1}{c|}{ViLT}& \multicolumn{1}{c|}{100\%}   & 83.40  & 85.86  & \multicolumn{1}{c|}{78.81}  & 72.96  & 77.57  & \multicolumn{1}{c|}{62.50} & 74.10 & -  & \multicolumn{1}{c|}{74.14}     & \multicolumn{1}{c}{68.85}&\\

\midrule
\rowcolor{gray!20}\textbf{Parameter-efficient Transfer Learning}   & &   &  &  &  &&  && &  & &   & &\\
\midrule

\multicolumn{1}{c|}{DARA~\cite{liu2024dara}}  & \multicolumn{1}{c|}{ICME'24} &\multicolumn{1}{c|}{RN50+DETR/BERT-B}& \multicolumn{1}{c|}{7.14\%}   & 81.16   & 82.76  & \multicolumn{1}{c|}{76.72}  & 65.58  & 69.83  & \multicolumn{1}{c|}{57.22} & 67.21 & 69.22 & \multicolumn{1}{c|}{67.67}    & \multicolumn{1}{c}{-}&\\

\multicolumn{1}{c|}{MaPPER\cite{liu2024mapper}}   & \multicolumn{1}{c|}{EMNLP'24} &\multicolumn{1}{c|}{DINOv2/BERT-B}& \multicolumn{1}{c|}{\underline{6.2\%}} & {86.03} & {88.90} & \multicolumn{1}{c|}{81.19} &  {74.92}& {81.12} & \multicolumn{1}{c|}{65.68}& {74.60} & {76.32} & \multicolumn{1}{c|}{75.81}&\multicolumn{1}{c}{-}& \\

\multicolumn{1}{c|}{HiVG\cite{xiao2024hivg}}   & \multicolumn{1}{c|}{MM'24} &\multicolumn{1}{c|}{CLIP-B}& \multicolumn{1}{c|}{23.04\%} & {87.32} & {89.86} & \multicolumn{1}{c|}{83.27} &  {\textbf{78.06}}& \textbf{84.81} & \multicolumn{1}{c|}{68.11}& {-} & {78.29} & \multicolumn{1}{c|}{78.79}&\multicolumn{1}{c}{82.11}& \\

\multicolumn{1}{c|}{\textbf{SwimVG (Ours)}}   & \multicolumn{1}{c|}{-}& \multicolumn{1}{c|}{DINOv2/CLIP-B}& \multicolumn{1}{c|}{\textbf{2.04\%}} & \textbf{88.29} & \textbf{90.37} & \multicolumn{1}{c|}{\textbf{84.89}} &  77.92& 83.22 & \multicolumn{1}{c|}{\textbf{69.95}}& \textbf{79.10} & \textbf{80.14} & \multicolumn{1}{c|}{\textbf{79.69}}  &\multicolumn{1}{c}{\textbf{83.10}} &\\
\bottomrule

\end{tabular}
}
\vspace{-3mm}

\end{table*}

\begin{equation}
\begin{aligned}
\begin{split}
    t_{down} = f_i^t\mathbf{W}_{down}, \\
    t_{act}=\operatorname{ReLU}\boldsymbol(t_{down}), \\
    t_{up} = t_{act}\mathbf{W}_{up},    
\end{split}
\end{aligned}
\end{equation}

\begin{equation}
\text{DoSA}(f_i^t)=f_i^t + s_t \cdot t_{up},
\end{equation}

\noindent
where $\mathbf{W}_{\text{down}} \in \mathbb{R}^{C_t \times C_d}$ and $\mathbf{W}_{\text{up}} \in \mathbb{R}^{C_d \times C_t}$ are the weights of down- and up-projection layers, and $s_t$ is the scaling factor of domain-specific adapters.
In this way, DoSA can refine the rich pre-trained language representations into more fine-grained representations for the VG domain during fine-tuning.

\subsection{Prediction Head}
Followed by HiVG \cite{xiao2024hivg} and TransVG++ \cite{transvg++}, a regression block with a MLP and a linear layer are adopted to perform box coordinates prediction. Given the [REG] token from the last layer of vision encoder, the regression block generates the 4-dim bounding box coordinates.

\subsection{Training Objectives}
Following the previous work \cite{deng2021transvg,liu2024dara}, the L1 loss and Generalized IoU (GIoU) loss are used between the the predicted bounding box coordinates $\tilde{b} = (\tilde{x}, \tilde{y}, \tilde{w}, \tilde{h})$ and the the ground truth ${b} = ({x}, {y}, {w}, {h})$, the training objective for VG is defined as follows:
\begin{equation}
\mathcal{L}_{\text{rec}} = \lambda_1 \mathcal{L}_{L1}(b, \tilde{b}) + \lambda_{\text{giou}} \mathcal{L}_{\text{giou}}(b, \tilde{b}),
\end{equation}

\noindent
where $\mathcal{L}_{L1}(\cdot)$ and $\mathcal{L}_{\text{giou}}(\cdot)$ represent L1 loss and GIoU loss \cite{rezatofighi2019generalized}], respectively. The $\lambda_1$ and $\lambda_{\text{giou}}$ are the weight coefficient to balance the two detection loss functions.

\section{Experiments}
In this section, we will give a detailed experimental analysis of the whole framework, including the datasets, evaluation protocol, implementation details, comparisons with the state-of-the-art methods, and ablation analysis.

\subsection{Experimental Setup}
\noindent \textbf{Datasets.}
To verify the effectiveness and efficiency of our method, we have conducted comprehensive experiments on the RefCOCO \cite{yu2016modeling}, RefCOCO+ \cite{yu2016modeling}, RefCOCOg \cite{mao2016refcocogg,nagaraja2016refcocogu} and Flickr30K Entities \cite{plummer2015flickr30k} datasets, all of which are widely used as benchmarks for visual grounding.

\begin{itemize}
  \item \textbf{RefCOCO} features 19,994 images with 50,000 referred objects and 142,210 expressions. The dataset is divided into four subsets, consisting of 120,624 train, 10,834 validation, 5,657 test A, and 5,095 test B samples, respectively. The average length of the expressions is 3.6 words, and each image contains a minimum of two objects.

 \item
\textbf{RefCOCO+} with similar content but richer expressions, includes 19,992 images with 49,856 referred objects and 141,564 referring expressions. The dataset is divided into four subsets: 120,624 train, 10,758 validation, 5,726 test A, and 4,889 test B samples. Notably, the RefCOCO+ dataset has been constructed to be more challenging than the RefCOCO dataset by excluding certain types of absolute-location words. The average length of the expressions is 3.5 words, including the attribute and location of referents.

 \item
\textbf{RefCOCOg} , unique for its detailed annotations and longer referential expressions, contains 25,799 images with 49,856 objects. There are two commonly used split protocols for this dataset. One is RefCOCOg-google \cite{mao2016refcocogg}, and the other is RefCOCOg-umd \cite{nagaraja2016refcocogu}. We report our performance on both RefCOCOg-google (val-g) and RefCOCOg-umd (val-u and test-u) to make comprehensive comparisons. The average length of expressions within the dataset is 8.4 words, including both the attributes and the locations of the referents. This rich detail description facilitates a more nuanced understanding of the visual grounding tasks, as it captures the intricacies of how objects are referenced in various contexts.

 \item
\textbf{Flickr30K Entities} \cite{plummer2015flickr30k}, is an enhanced version of the original Flickr30K \cite{young2014image}, fortified with the addition of short region phrase correspondence annotations. This expansion yields a collection of 31,783 images, encompassing 427,000 referred entities. Following the previous studies \cite{xiao2024hivg,wang2023cogvlm}, we have divided the dataset into 29,783 images for training, 1,000 for validation, and another 1,000 for testing purposes.
\end{itemize}

\noindent
\textbf{Evaluation Metrics.} We follow the previous research that employs top-1 accuracy (\%) as the evaluation metric for visual grounding. Specifically, a prediction is deemed accurate only when its Intersection-over-Union (IoU) exceeds or equals 0.5. In addition to Precision@0.5, we also report the number of tunable parameters in the pre-trained encoders to compare the fine-tuning efficiency with traditional full fine-tuning and other PETL methods.

\begin{table*}[!t]
\centering
\caption{Comparison with PETL methods using the same Backbone as SwimVG on RefCOCO, RefCOCO+ and RefCOCOg. ``Param.'' indicates the number of tunable parameters in the pre-trained encoders.}
\vspace{-3mm}
\label{Table:comparisons with PETL}
\small
\setlength{\tabcolsep}{4.5pt}

\begin{tabular}{l|c|ccc|ccc|ccc}
\toprule
    
\multirow{2}{*}{Methods} &\multicolumn{1}{c|}{\multirow{2}{*}{Venue}}  & \multicolumn{3}{c|}{RefCOCO} & \multicolumn{3}{c|}{RefCOCO+} & \multicolumn{3}{c}{RefCOCOg} \\

 && val & testA & testB & val & testA & testB & val-g & val-u & test-u \\ \midrule

AdaptFormer \cite{chen2022adaptformer} &NeurIPS'22 & 81.75 & 83.14 & 76.73 & 72.05 & 76.61 & 64.26 & 70.19 & 70.93 & 72.36 \\

LoRA \cite{hu2021lora}  & ICLR'22  & 82.43 & 84.51 & 77.32 & 72.66 & 77.13 & 64.85 & 71.27 & 72.16 & 73.23 \\

UniAdapter \cite{lu2024uniadapter}& ICLR'24 & 85.76 & 88.31 & 81.84 & 74.95 & 78.75 & 65.97 & 73.68 & 74.72 & 74.98 \\

DAPT \cite{zhou2024dynamic} &CVPR'24 & 85.33  & 87.52 & 81.06 & 74.33 & 78.66& 65.54 & 74.02 & 75.26 & 75.47   \\

\midrule
\textbf{SwimVG}  &-& \textbf{88.29} & \textbf{90.37} & \multicolumn{1}{c|}{\textbf{84.89}} &  \textbf{77.92}& \textbf{83.22} & \multicolumn{1}{c|}{\textbf{69.95}}& \textbf{79.10} & \textbf{80.14} & \textbf{79.69}  \\
\bottomrule

\end{tabular}

\end{table*}

\noindent \textbf{Implementation Details.} The vision encoder is initialized with DINOv2-L/14~\cite{oquab2023dinov2}, while the language encoder uses CLIP-B~\cite{radford2021learning}. The resolution of the input image is 224×224. The DINOv2-L/14 model processes tokens with a feature dimension of 768, while 
and the CLIP-B model handles tokens with a feature dimension of 512. All prompts use Xavier initialization, and all adapters are initialized with Kaiming normal initialization. The bottleneck dimension $C_d$ for both CIA and domain-specific adapters is 56, and more dimension comparisons can be seen in Table \ref{Table:neck}. The batchsize for training is 32. For fair comparisons, other PETL methods in Tab. \ref{Table:comparisons with PETL} use the same base architecture and original hyperparameters, and keeping the vision and language encoder frozen. For RefCOCO \cite{yu2016refcoco}, RefCOCOg \cite{mao2016refcocogg,nagaraja2016refcocogu}, and Flickr30K Entities \cite{plummer2015flickr30k} datasets, the entire network is trained for 65 epochs using the AdamW optimizer. While for RefCOCO+ \cite{yu2016refcoco} dataset, the network is trained for 90 epochs. Note that most mainstream methods train RefCOCO/RefCOCOg/Flickr30K Entities for 90 epochs and RefCOCO+ for 180 epochs, which demonstrates the higher efficiency of our SwimVG. We conduct all experiments on one A800 GPU.

\subsection{Main Results}
We compare our SwimVG comprehensively with a series of previous visual grounding (VG) methods. The main experimental results are displayed in Tab. \ref{Table:comparisons with SOTA}. We can notice from these results that SwimVG reaches the best accuracy and also ensures parameter efficiency compared with all other methods, which validates its effectiveness and efficiency.

\noindent
\textbf{Effectiveness.} As Tab. \ref{Table:comparisons with SOTA} shown, on the three commonly challenging benchmarks, SwimVG outperforms all traditional full fine-tuning methods. Compared to DARA~\cite{liu2024dara}, a parameter-efficient transfer learning method, we achieves an average accuracy improvement of 10.85\% on the three benchmarks. Notably, even compared to some methods that are pre-trained on the the RefCOCO/+/g and Flickr30K Entities (indicated by $\dagger$ in Tab. \ref{Table:comparisons with SOTA}), our SwimVG model achieves the highest scores across all evaluation tasks, with particularly strong performance on the RefCOCO+, which present greater challenges compared to RefCOCO.

\noindent
\textbf{Efficiency.} Tab. \ref{Table:comparisons with SOTA} clearly illustrates that SwimVG not only achieves the best performance, but also highlights its huge advantages in parameter efficiency. SwimVG reduced the tunable backbone parameters by 97.96\% compared to the traditional full fine tuning method. In order to verify more efficient aspects such as training and inference time, experimental results on the mainstream methods using the conventional VL transformer, and the other PETL methods are shown in Tab. \ref{tab:cost}. It can be seen that SwimVG achieves significant energy efficiency advantages.

\begin{table}\footnotesize

\begin{center}
\renewcommand{\arraystretch}{1.2}
\caption{Efficiency comparison. The results are obtained on RefCOCO dataset. ``$-$'' indicates that the model's code is not publicly available, and their results are not available.}
\label{tab:cost}
\resizebox{1.0\columnwidth}{!}{%
    \begin{tabular}{c|c|c|c|c|c}
    \toprule
    \multirow{2}{*}{Model} & update/all  & update        & train time                & testA time          & testA  \\
                         
                           &  param.     & ratio  & (epoch/min)        & (s)   & Acc.$\uparrow$  \\
                          
    \midrule
 \textbf{Full Fine-tuning}   & \multicolumn{1}{c}{} & \multicolumn{1}{c}{} &  \multicolumn{1}{c}{}  & \multicolumn{1}{c}{} &  \\
\midrule
    TransVG          & 159.4/159.4M   & 100$\%$   & 52        & \underline{95}   & 82.67 \\
    QRNet            & 281.4/281.4M   & 100$\%$    &62       & 111  & 85.85 \\ 
    VG-LAW    & 158.7/158.7M   & 100$\%$   & --           & --    & 88.56 \\   
    TransVG++ & 171.13/171.13M   & 100$\%$   & --            & --    & 88.37 \\  
    \midrule
 \textbf{PETL Methods}   & \multicolumn{1}{c}{} & \multicolumn{1}{c}{} & \multicolumn{1}{c}{}   & \multicolumn{1}{c}{} &  \\
 \midrule
    DARA & 11.61/162.61M   & 7.14$\%$    &           \textbf{33} &  96  & 82.76 \\  
    LoRA & 17.15/392.15M   & 4.37$\%$   & 61          & 127    & 84.51  \\  
    AdaptFormer & 14.85/389.85M   & \underline{3.81$\%$}    &         57  &125    &  83.14  \\  
    Uniadapter & 29.16/404.16M   & 7.21$\%$    &          65 & 131   & 88.31  \\  
    DAPT & 26.69/401.69M   & 6.64$\%$    &           64  & 129    &87.52 \\  
    HiVG & 49.40/214.40M   & 23.04$\%$    &          - & -   & \underline{89.86} \\  

    \midrule
    \textbf{SwimVG}(ours)  & {7.65/375.13M}    & \textbf{2.04$\%$}  & {\underline{40}}       & {\textbf{65}}  & \textbf{90.37} \\  
    \bottomrule
\end{tabular}%
}
\end{center}

\vspace{-11pt}	
\end{table}

\subsection{Comparison with Other PETL Methods}

\textbf{Details of Baseline PETL Methods.}

This section furnishes additional details of the PETL baselines employed in our primary manuscript. Notably, all these baselines follow the same base architecture.
\begin{itemize}

    \item \textbf{AdaptFormer \cite{chen2022adaptformer}:} We add adapters in parallel to MHA and FFN in both Vision Encoder and Language Encoder. Following the original work, we set the same bottleneck dimensions of AdaptFormer for both vision and language branch.
    
    \item \textbf{LoRA \cite{hu2021lora}:} We incorporate trainable matrices in parallel to the weight matrices in MHA and FFN in both Vision Encoder and Language Encoder. 
    We have set the same bottleneck dimensions for both the vision and language branches of LoRA, following the original setup.

    \item \textbf{UniAdapter \cite{lu2024uniadapter}:} We add UniAdapter  in both Vision Encoder and Language Encoder, according to their basic designs. 

    \item \textbf{DAPT \cite{zhou2024dynamic}:} We insert Dynamic Adapters in paralle to the weight matrices in MHA and FFN in both Vision Encoder and Language Encoder, and use their task-agnostic feature transform strategy. Other sets such as bottleneck dimensions are same as the DAPT.

    
\end{itemize}

We conduct experiments comparing our SwimVG with other parameter-efficient transfer learning (PETL) methods. To ensure fairness, we retain the original parameter settings from previous methods. As these PETL methods lack the capability of multimodal fusion, we complement them with the traditional VL transformer for cross-modal understanding, thereby enabling a direct comparison with our SwimVG. Tab. \ref{Table:comparisons with PETL} illustrates that SwimVG outperforms other PETL methods on all three benchmarks. Through introducing step-wise multimodal prompts and cross-modal interactive adapters, SwimVG enhances the modeling of the vision-text alignment capability. Previous PETL methods lack this ability, rendering them less effective for VG tasks. This also proves that the multimodal fusion mechanism in SwimVG is more efficient than the VL transformer. To summarize, by the specific design for the VG domain, SwimVG achieves superior performance with only \textbf{2.04} \% tunable parameters.

\begin{figure*}[t]
\centering
\includegraphics[width=0.98\textwidth]{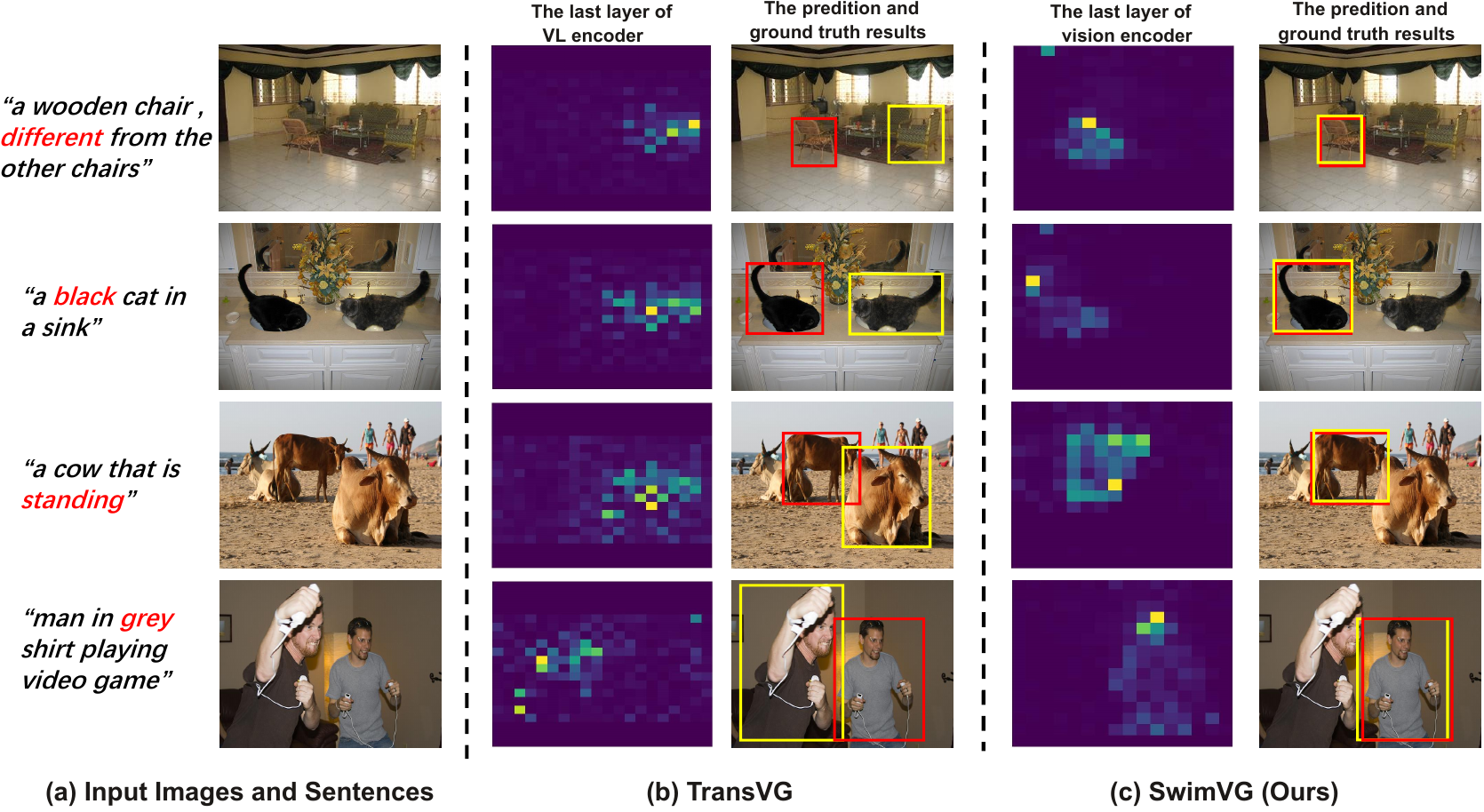}
\caption{Visualizations of attention maps, prediction results (yellow bounding boxes) and ground truth (red bounding boxes).}
\label{fig:visualizations}
\end{figure*}

\subsection{Convergence Analysis}
\begin{figure}
\centering
\includegraphics[width=1\linewidth]{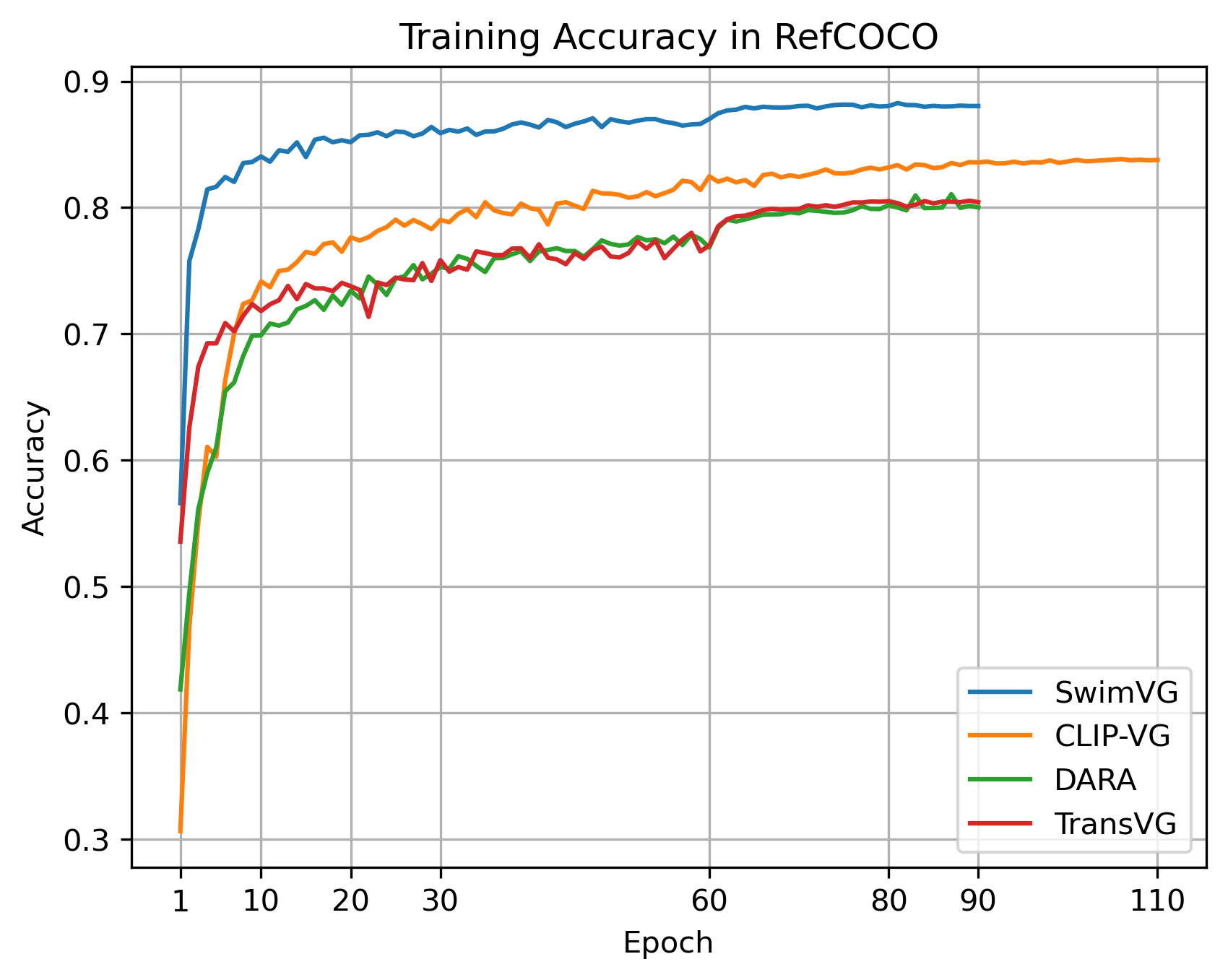}
\vspace{-4mm}
\caption{The convergence comparison between SwimVG and other SOTA models on RefCOCO.}
\vspace{-3mm}
\label{other-convergence}
\end{figure}

\begin{figure}
\centering
\includegraphics[width=1\linewidth]{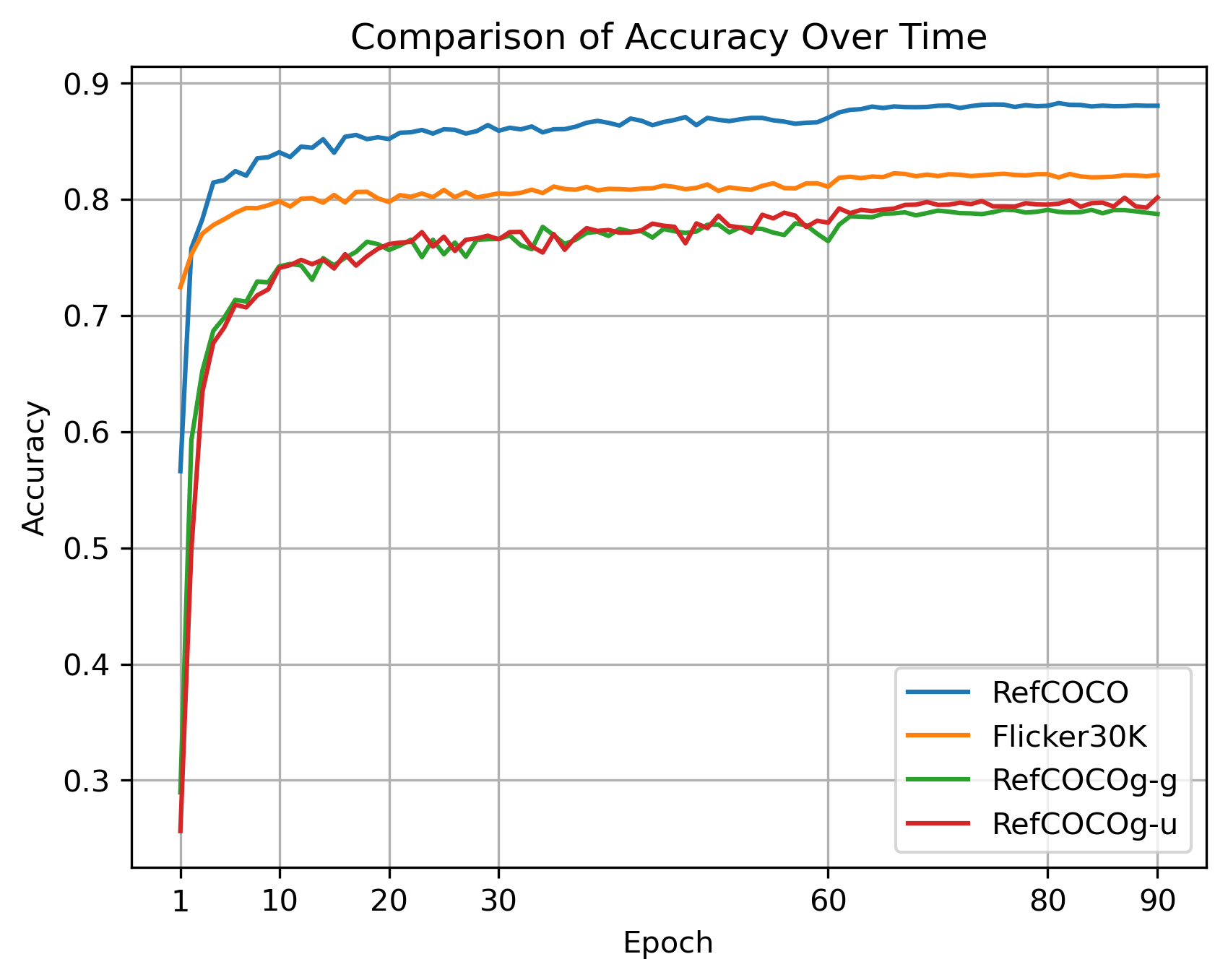}
\vspace{-4mm}
\caption{The convergence comparison of SwimVG on RefCOCO, RefCOCOg and Flicker 30K datasets.}
\vspace{-3mm}
\label{self-conver}
\end{figure}

Figure \ref{other-convergence} shows a comparison of the convergence epoch between SwimVG and other models. It is observed that DARA and TransVG converge around epoch 85, while CLIP-VG converges at approximately epoch 105. In contrast, SwimVG achieves convergence at around epoch 65. This demonstrates the efficiency of our method, as fewer training epochs are required, thereby reducing training costs. In addtion, we have also visualized the convergence comparison of SwimVG across the RefCOCO, RefCOCOg-u, RefCOCOg-g, and Flicker 30K datasets. Figure \ref{self-conver} indicates that convergence is achieved around epoch 65 for all these datasets.

\subsection{Ablation Study}

\noindent
\textbf{Effectiveness of Multimodal Interaction in SwimVG.} We assess the impact of step-wise multimodal prompts (Swip) and cross-modal interactive adapters (CIA) by performing an ablation study, and report the results on RefCOCOg-u validation and test datasets. Considering the substantial number of parameters occupied by the encoder, we freeze all the encoder parameters during fine-tuning for efficiency. From Tab. \ref{Table:swip-cia}, it is evident that only introducing the Swip yields a ideal results (Tab. \ref{Table:swip-cia} (a)). Only by using the CIA for cross-modal fusion can achieve better results (Tab. \ref{Table:swip-cia} (b)). Compared with the previous methods using the traditional vision-language encoder, such as TransVG \cite{deng2021transvg}, DARA \cite{liu2024dara} in Tab. \ref{Table:comparisons with SOTA}, it shows that we can achieve the better results using only Swip or CIA. Tab. \ref{Table:swip-cia} (c) indicates that incorporating Swip and CIA for multimodal fusion results in an average improvement of 3.49\% across the RefCOCOg-u, achieving the best performance among these ablation variants. Swip achieves progressive multimodal fusion by gradually introducing linguistic information, while CIA explores deeper correlations by enhancing cross-modal interaction. Combining the two can simultaneously promote multimodal fusion in terms of breadth and depth.

\begin{table}[t]
\centering
\renewcommand{\arraystretch}{1.5}
\caption{Ablations of multimodal interaction in SwimVG on RefCOCOg-u \cite{yu2016modeling} dataset. Note that the visual and text encoder are frozen in the ablation studies.}
\vspace{-2mm}
\label{Table:swip-cia}
\small
\setlength{\tabcolsep}{4pt}
\begin{tabular}{ccc|ccc}
\toprule
& Step-wise & Cross-modal  & Updated  & \multicolumn{2}{c}{RefCOCOg} \\
 & Multi. Prompts & Inter. Adapters  &  Params. & val-u & test-u \\ \midrule

(a)  & $\checkmark$ &    & 6.30M & 71.32  &70.06 \\
(b)  &  & $\checkmark$  &1.00M&72.22 & 71.86\\
  
(c)  &$\checkmark$ & $\checkmark$  &7.30M & \textbf{75.57} & \textbf{75.48} \\ \bottomrule
\end{tabular}

\end{table}

\noindent
\textbf{Effectiveness of Domain-specific Adapters.} Because the text encoder is pre-trained on a general domain, freezing the entire text backbone restricts the specific language understanding in visual grounding domain, thereby weakening the proper interaction between text and vision semantics. To enable the domain text semantics to interact with the visual encoder efficiently, we adopt domain-specific adapters to learn the domain knowledge, thus making the text encoder match with visual grounding. Tab. \ref{Table:ablation on text adapter} shows that domain-specific adapters efficiently transfer the language knowledge of the pre-trained model to VG domain, further improving an average improvement of 4.39\% across the RefCOCOg-u.

\begin{table}[!t]
\centering
\renewcommand{\arraystretch}{1.5}
\caption{Effectiveness of Domain-specific adapters. (a) represents introducing Swip and CIA in SwimVG.}
\vspace{-2mm}
\label{Table:ablation on text adapter}
\small
\setlength{\tabcolsep}{10pt}

\begin{tabular}{l|c|c|ccc}
\toprule
    
 \multirow{2}{*}{\#} & {Domain-specific } & \multicolumn{1}{c|}{Updated} & \multicolumn{2}{c}{RefCOCOg} \\

 & Adapters& Params. & val-u & test-u  \\ \midrule

(a)  &  & 7.30M &  75.57 & 75.48  \\ 
\midrule
(b) & $\checkmark$ & 7.65M  &  \textbf{80.14}& \textbf{79.69}  \\

\bottomrule

\end{tabular}

\end{table}

\begin{figure*}[t]
\centering
\includegraphics[width=\textwidth]{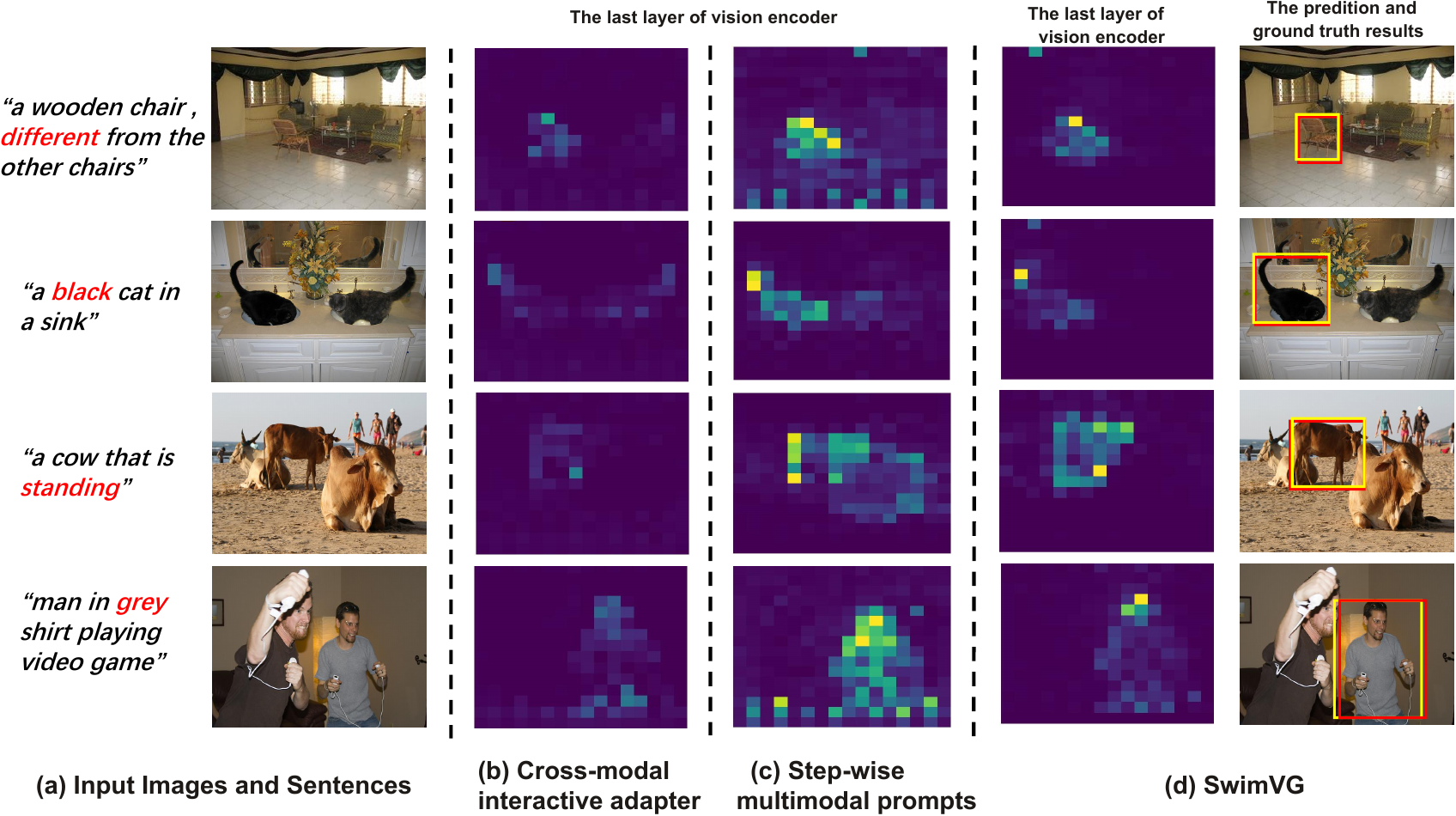}
\caption{The visualizations of attention maps from vision encoder with different strategies of SwimVG. Red bounding boxes represent ground truth, and yellow bounding boxes are prediction results.}
\vspace{-4mm}
\label{fig:more visualizations}
\end{figure*}

\noindent
\textbf{Effects of Different Insertion Positions of SwimVG.} To determine the optimal configuration of the Cross-modal Interactive Adapter (CIA) and Text Adapter, we conducted an ablation study varying both different layers and the dimensions of the adapters. Firstly, we evaluated the impact of different adapter layers. In this experiment, the visual CIA and the Text Adapter were inserted at the same layers. From Table \ref{tab:layer}, we can see that: \textbf{(1)} Only inserting three layers for vision and text encoder can brings great performance (Table \ref{tab:layer} (a)); \textbf{(2)} observing Table \ref{tab:layer} (b), (c), and (d), it can be seen that inserting CIA later in the vision encoder can exhibit better performance; \textbf{(3)} 
from the observation of Table \ref{tab:layer} (e) and (f), it is evident that inserting text adapter later in the text encoder results in a minor performance decline; \textbf{(4)} 
adding adapters from 13 layers to 24 layers not only reduces performance but also increases the tunable parameters. This might be because the visual backbone is more likely to adapt to the VG domain at deeper layers, while the text needs to adapt from the shallow layers to the deep layers. It should be noted that the text encoder is composed of 12 layers, while the vision encoder comprises 24 layers.

\begin{table}[t]
\centering
\renewcommand{\arraystretch}{1.5}
\caption{Ablation study of different configurations of cross-modal interactive adapters and text adapters. For the ``Position'', we list the i-th layers that insert adapters in the backbone.
}
\label{tab:layer}
\small
\setlength{\tabcolsep}{3pt}
\begin{tabular}{l|l|l|c|lll}

\toprule
\multirow{2}{*}{\#} & \multicolumn{2}{c|}{Position} & \multirow{2}{*}{Params}  & \multicolumn{2}{c}{RefCOCOg} \\ \cline{7-7} 
 &  \multicolumn{1}{c|}{text}  &  \multicolumn{1}{c|}{vision} & & \multicolumn{1}{c|}{val-u} & \multicolumn{1}{c}{test-u}  \\ \midrule
(a) &4,8,12 & 8,16,24 & 6.67M   & \multicolumn{1}{l|}{75.26} & 74.78 \\

(b) &2,4,6,8,10,12 & 4,8,12,16,20,24 & 7.65M   & \multicolumn{1}{l|}{78.65} & 72.54 \\ 
 
(c) &2,4,6,8,10,12 & 14,16,18,20,22,24 &7.65 M  & \multicolumn{1}{l|}{79.28} & 78.62 \\

\rowcolor{gray!20}
(d) & 2,4,6,8,10,12& 19,20,21,22,23,24 & 7.65M  & \multicolumn{1}{l|}{80.14} & 79.69 \\

(e) &7,8,9,10,11,12& 19,20,21,22,23,24 & 7.65M   & \multicolumn{1}{l|}{78.90} & 78.06 \\

(f)  & 7,8,9,10,11,12& 14,16,18,20,22,24 & 7.65M  & \multicolumn{1}{l|}{79.06} & 78.43 \\

(g) & 2,4,6,8,10,12& 13-24 & 8.65M  & \multicolumn{1}{l|}{79.51} & 78.39 \\

\bottomrule
\end{tabular}

\end{table}

\begin{table}[!t]
\centering
\renewcommand{\arraystretch}{1.5}
\caption{Effectiveness of different bottleneck for all adapters.}
\label{Table:neck}
\small
\setlength{\tabcolsep}{8pt}

\begin{tabular}{l|c|c|ccc}
\toprule
    
 \multirow{2}{*}{\#} & \multirow{2}{*}{Bottleneck dimensions} & \multicolumn{1}{c|}{Params.} & \multicolumn{2}{c}{RefCOCOg} \\

 & & (M) & val-u & test-u  \\ \midrule
(a)  &32 & 7.05 &  78.65 & 78.13  \\
(b)  &40 & 7.24 &  79.67 & 78.78  \\
\rowcolor{gray!20}
(c) & 56&7.65  & 80.14 &79.69  \\
(d) & 64& 7.87  & 79.12 & 78.52 \\
(e) & 128&  9.76 & 80.18 &79.43  \\

\bottomrule

\end{tabular}
\vspace{-2mm}

\end{table}

\noindent
\textbf{Effects of Different Hyper-parameter Settings of SwimVG.} We first ablate the bottleneck dimensions $C_d$ of all adapters (see Table \ref{Table:neck} (a,b,c)), and follow the design shown in Table \ref{Table:neck} (a). $C_d$ determines the number of tunable parameters introduced by SwimVG. As shown in Table \ref{Table:neck}, higher $C_d$ introduces more parameters, and the performance consistently increases when $C_d$ increases up to 56. $C_d$ 128 exhibits considerable performance, but its tunable parameter count is about twice that of $C_d$ 56. Thus, we select the $C_d$ as 56. This indicates that a small bottleneck may not provide sufficient adaptation capabilities, while a large dimension may lead to over-adaptation. An intermediate dimension can achieve a better adaptation to the VG domain.

\begin{table}[t]
\caption{Comparison of the contribution levels of different backbones.}
\centering

\small
\setlength{\tabcolsep}{3.5pt}

\begin{tabular}{l|cc|ccc}
\toprule

\multirow{2}{*}{Mehthods} & \multicolumn{1}{c}{Vision} & \multicolumn{1}{c|}{Language} & \multicolumn{3}{c}{RefCOCO} \\

 & Backbone & Backbone & val & testA & testB \\ \midrule
TransVG\cite{deng2021transvg} &RN101+DETR & BERT-Base & 81.02 & 82.72 & 78.35
\\
TransVG\cite{deng2021transvg} & DINOv2-L & BERT-Base  & 85.11 & 87.36  & 80.97
\\

TransVG\cite{deng2021transvg} & DINOv2-L & CLIP-Base & 85.55 & 86.79  & 80.28
\\
\hline
SwimVG & DINOv2-L & CLIP-Base & \textbf{88.29}  & \textbf{90.37 }&\textbf{84.89}
\\
\bottomrule
\end{tabular}

\vspace{-2mm}
\label{Table:backbone}
\end{table}
\noindent
\textbf{The contribution degree of different pre-trained models.} To facilitate the analysis of the contribution of different backbones to performance, we excluded the SwimVG method and compared different backbones based on TransVG\cite{deng2021transvg}. We selected ResNet101+DETR and DINOv2-L as the vision backbone and chose the mainstream BERT-Base and the text encoder in CLIP-Base as the text backbone. As see in Table \ref{Table:backbone}, the vision backbone has a relatively large impact on visual grounding, whereas the text backbones have a relatively small impact. Under the same backbone, our method outperforms TransVG, which indicates that our multimodal fusion strategy is highly effective.

\vspace{-4mm}
\subsection{More Evaluation Metrics}
We compared more challenging evaluation metrics, such as the prediction accuracy when IoU $>$ 0.6 (Pr@0.6) and Pr@0.8. Under the same metrics, we compared the latest MaPPER \cite{liu2024mapper}. As seen in Table \ref{Table:iou}, SwimVG outperforms the latest MaPPER under both the settings of Pr@0.6 and Pr@0.8.

\begin{table}[t]
\caption{Comparison of the more evaluation metrics.}
\centering

\small
\setlength{\tabcolsep}{5pt}

\begin{tabular}{l|ccc|ccc}
\toprule

\multirow{2}{*}{Mehthods} & \multicolumn{3}{c|}{Pr@0.6(RefCOCO)} & \multicolumn{3}{c}{Pr@0.8(RefCOCO)} \\

& val & testA & testB & val & testA & testB \\ \midrule
MaPPER\cite{liu2024mapper} & 82.23 & 86.03 & 76.11 & 66.62 & 72.63 &57.50
\\
SwimVG & \textbf{85.26} & \textbf{87.33} & \textbf{80.61}  & \textbf{68.86} & \textbf{72.83} & \textbf{63.04}
\\
\bottomrule
\end{tabular}

\vspace{-2mm}
\label{Table:iou}
\end{table}

\subsection{Qualitative Results}
\textbf{The comparison of multimodal fusion strategy.}
To verify that the multimodal fusion strategy of SwimVG is superior to the traditional vision-language transformer (VL encoder), we visualize the attention maps from the last layer of vision encoder in SwimVG. Due to the suboptimal multimodal fusion methods employed by other mainstream approaches, namely the visual language transformer (VL encoder), which lack open-source code or checkpoints, we opt to visualize the last layer of the VL encoder from TransVG. As shown in Fig.\ref{fig:visualizations}, TransVG fails to pay sufficient attention to text-relevant regions in a images. For example, TransVG lacks the alignment ability of ``$different$'', ``$black$'', and ``$standing$'' with images. The comparison with TransVG demonstrates the ability of our proposed SwimVG to focus more on the text-relevant regions, and our multimodal fusion strategy is superior to the traditional VL encoder.

\noindent
\textbf{The effectivess of CIA and Swip.}
In this section, we present more visualization of the attention maps from the vision encoder under different mixing strategies. As depicted in Figure \ref{fig:more visualizations}, we can see that: \textbf{(1)} introducing either cross-modal interactive adapters (CIA) or step-wise multimodal prompts (Swip) facilitates the interaction between the vision and language encoders. (Figure \ref{fig:more visualizations} (b,c)); \textbf{(2)} compared to CIA, the attention map of only introducing is slightly scattered (Figure \ref{fig:more visualizations} (b,c)); 
integrating CIA and Swip can further enhances the facilitation of cross-modal interaction (Figure \ref{fig:more visualizations} (d)). The interaction between the vision and language encoder, facilitated by CIA and Swip, allows the model to focus more effectively on the referred objects in diverse expression cases.

\section{Conclusion and Future Work}
\subsection{Conclusion}
In this paper, we aims at improving both the effectiveness and efficiency of visual-text alignment. We propose SwimVG by the foundational design of step-wise multimodal prompts (Swip) and cross-modal interactive adapters (CIA). SwimVG integrates a novel multimodal fusion strategy of token-level Swip and weight-level CIA to enable the visual encoder can concentrate on the text-relevant regions. Extensive experiments and ablation studies have validated the high effectiveness of our method. Our proposed framework significantly outperforms the baseline and achieves comparable results with the state-of-the-art methods while tiny parameter budget.

\subsection{Future Work}

In the future, implementing our SwimVG in real-world applications is a challenging and meaningful direction. Currently, our SwimVG has only been evaluated on benchmark datasets. However, its performance against datasets from different domains remains unknown.
In addition, the efficient multi-modal fusion strategies of SwimVG can be verified on other multimodal tasks, such as visual question answering and video caption. Motivated by efficient Multimodal Large Language Model \cite{liu2024multi}, we will explore efficient training and inference model for visual grounding.


{
\bibliographystyle{IEEEtran}
\bibliography{ref}
}

\end{document}